\documentclass{article}

    \PassOptionsToPackage{numbers, compress}{natbib}

    \usepackage[preprint]{neurips_2020}

\usepackage[utf8]{inputenc} 
\usepackage[T1]{fontenc}    
\usepackage{hyperref}       
\usepackage{url}            
\usepackage{booktabs}       
\usepackage{amsfonts}       
\usepackage{nicefrac}       
\usepackage{microtype}      
\usepackage{natbib}
\usepackage{tablefootnote}

\title{Teach me how to Label: Labeling Functions from  Natural Language with Text-to-text Transformers}

\author{%
  Yannis Papanikolaou \\
  \\
  \texttt{yanpapanikolaou@gmail.com} \\

}

\begin{document}

\maketitle

\begin{abstract}
Annotated data has become the most important bottleneck in training accurate machine learning models, especially for areas that require domain expertise. A recent approach to deal with the above issue proposes using natural language explanations instead of labeling individual data points, thereby increasing human annotators' efficiency as well as decreasing costs substantially. This paper, focuses on the task of turning these natural language descriptions into Python labeling functions by following a novel approach to semantic parsing with pre-trained text-to-text Transformers. In a series of experiments our approach achieves a new state of the art on the semantic parsing benchmark CoNaLa, surpassing the previous best approach by 3.7 BLEU points. Furthermore, on a manually constructed dataset of natural language descriptions--labeling functions pairs we achieve a BLEU of 0.39. Our approach can be regarded as a stepping stone towards models that are taught how to label in natural language, instead of being provided specific labeled samples. Our code, constructed dataset and models are available at \url{https://github.com/ypapanik/t5-for-code-generation}
\end{abstract}

\section{Introduction}
Recent advances in the field of deep learning \citep{szegedy2015going, he2016deep, devlin2018bert, raffel2019exploring} have resulted in models that achieve human-like performance in a series of challenging benchmarks. Nevertheless, transferring these advances to real-world tasks are most often hindered by the lack of annotated data. 

To obtain labeled data, the classic approach relies on humans to provide gold annotations. This approach becomes quickly inefficient and expensive, especially for domains that require annotator expertise such as biomedicine. These limitations became very quickly evident to researchers, who tried a number of approaches to decrease the need for human annotation. Distant supervision approaches \citep{mintz2009distant} make use of structured knowledge (e.g., from knowledge bases) to create automatically noisy annotated data. Active learning techniques \citep{settles2009active} attempt to find instances that will contribute the most in improving a model and subsequently ask humans for annotation of these specific instances, thereby reducing the cost of labeling. More recently, weak supervision \citep{ratner2016data}, has enabled the use of noisy or imprecise sources as a high-level supervision signal for labeling multiple data points, whereas large pretrained models \citep{devlin2018bert, liu2019roberta, raffel2019exploring} have allowed for robust transfer learning across domains needing fewer gold data.

One of the most promising aforementioned approaches, weak supervision \citep{ratner2016data}, shifts the role of human annotators from annotating specific data points to providing Labeling Functions (LFs) which are then applied on the data, providing large, noisy training sets. In this setting, LFs are then combined through a probabilistic framework to aggregate different sources of information and cancel out noise. This approach holds great promise in changing how data labeling is done, nevertheless it requires annotators to code directly LFs. 

In this paper, we consider the above setting and employ a semantic parsing approach \citep{yao2018staqc, yin2018mining, xu2020incorporating} to turn Natural Language (NL) descriptions into LFs. To the best of our knowledge, our work is the first to employ text-to-text Transformers \citep{raffel2019exploring} to the task of semantic parsing and by extension to convert NL explanations to LFs. Importantly, our approach results in a new state of the art in the semantic parsing benchmark CoNaLa \citep{yin2018mining} whereas it scores a BLEU of 0.39 on a manually labeled set of 193 NL to LF pairs. By improving semantic parsing of NL explanations, our work can be considered as a stepping stone towards models that learn directly from NL explanations and consequently need substantially less human supervision to solve machine learning tasks.

In the following, we present the relevant literature in Section \ref{sec:related_work}, our approach in Section \ref{sec:approach}, our experiments in Section \ref{sec:experiments} followed by the conclusions in Section \ref{sec:conclusions}.

\section{Related Work}
\label{sec:related_work}
A number of recent works has focused on learning from NL explanations. \citet{srivastava-etal-2017-joint} use semantic parsing and subsequently employ the resulting logical forms as features in a linear classifier, essentially using a traditional supervision approach with user-specified features. \citet{hancock2018training} develop a grammar-based semantic parser to turn NL descriptions to LFs for weakly supervising the creation of a larger training set via data programming.

More recently, A number of works \citep{ye2020teaching, zhou2020nero, wang2019learning, TriggerNER2020} have employed a similar approach across a variety of NLP tasks, semantic parsing having in all cases a critical role in converting NL explanations to executable logical forms.

Finally, although not directly related to our work, we would like to note two recent works that contribute interesting approaches to using NL explanations for solving machine learning tasks. \citet{murty2020expbert} first finetune BERT on a generic entailment task and subsequently provide to the finetuned model the input sequence as well as the relevant NL explanation, achieving equivalent performance to a BERT baseline with 3–20 times less labeled data.
\citet{liang2020alice} present a framework for computer vision tasks by using active learning to select the most informative pairs of label classes and then receiving contrastive NL explanations from annotators. Using a semantic parser they convert these explanations to LFs and incorporate the extracted knowledge through dynamically changing the learning model’s structure. Importantly, they show empirically that adding 1 explanation leads to similar performance gain as adding 13-30 labeled training data points.

Central to most of the aforementioned works is the task of semantic parsing. Crucially, an accurate semantic parser will also allow for "looser" human descriptions and allow for more extensive and informative NL explanations. We therefore view our work here as fundamental in allowing models to understand better the information provided by humans and maximize the benefit of these approaches.

\section{Approach}
\label{sec:approach}
Formally, given an NL description $x$, we are interested in generating a code snippet $c$ in Python, conditioned on $x$. Both $x$ and $c$ are sequences of text, therefore a natural choice is to consider a pretrained sequence-to-sequence Transformer language model and more specifically T5 \citep{raffel2019exploring}. T5 is a deep network architecture of  encoder-decoder Transformers \citep{vaswani2017attention} that has been pre-trained on both unsupervised and supervised tasks in a multi-task fashion. Unsupervised training is carried out with a variation of the “masked language modeling” (MLM) objective \citep{devlin2018bert} on a large collection of web crawled text, the "Colossal Clean Crawled Corpus", while for the supervised tasks the authors have used a mixture of benchmark datasets, such as GLUE\citep{wang-etal-2018-glue} and SQUAD\footnote{https://rajpurkar.github.io/SQuAD-explorer/}.

Even if the model has been pretrained on significantly different data and tasks, past work has shown \citep{devlin2018bert, raffel2019exploring, liu2019roberta} that Transformer-based pretrained language models can provide strong contextualized representations of language. Therefore, we hypothesize that finetuning T5 on a dataset that pairs NL descriptions with code snippets will allow the model to learn the specific language structure as well as attending correctly to word tokens in order to generate code elements, in a fashion similar to the task of machine translation. To finetune our model, we employ a standard sequence-to-sequence cross-entropy objective. In the following, we describe the experimental setup.

\section{Experiments}
\label{sec:experiments}
In this Section we describe the setup as well as the results of our experiments.

\begin{table}
\caption{\label{tbl:conala} Results on the CoNaLa benchmark. For other methods, results are taken from the relevant papers, while for the T5 models we report mean across five runs. \citep{xu2020incorporating} is the current state of the art.}

\centering
\begin{tabular}{ccc}
\hline 
Model&BLEU&Accuracy \\ 
\hline
\hline
Best submission\tablefootnote{https://competitions.codalab.org/competitions/19175}&14.72&-\\ 
\citet{yin2019reranking}&30.11&3.0 \\
\citet{xu2020incorporating}&32.26&- \\
\hline
T5-small&28.68&2.2\\
T5-base&31.34&3.6\\
T5-large&32.25&3.8\\
T5-small+noisy&32.52&4.6\\
T5-base+noisy&34.29&5.0\\
T5-large+noisy&\textbf{35.92}&\textbf{6.2}\\
\hline 
\hline 
\end{tabular}
\end{table}

\paragraph{Experiments on CoNaLa}
\label{sec:exp1}
In the first experiment, our goal is to understand how well can a T5 model learn to map NL to code. For this reason we employ the CoNaLa benchmark \citep{yin2018mining} which contains both a gold data set as well as a noisy set of 500k examples. We use the same split for gold train and test as provided by the authors and finetune T5 models on two different regimes: In the first case we use solely the gold training set provided, while in the second case we interleave batches from either the noisy set or the gold training set. To determine the optimal parameters, we use a subset of 200 examples from the training set as a development set. We use a batch size of 32 and a maximum sequence length of 32 for all models, while for the learning rate we used 0.001, 0.0005 and 0.0001 for T5-small, T5-base and T5-large respectively, finetuning for up to 30 epochs and keeping the best model on the development set.

In Table \ref{tbl:conala} we illustrate the results. As we can see, all T5 models  outperform the current state-of-the-art \citep{xu2020incorporating} even-though the latter incorporates external knowledge. These results are particularly important if one considers that T5 as well as its tokenizer haven't been explicitly trained on Python code data. Also, we observe that the noisy dataset allows the models to better understand the task and contributes significantly in improving results.

\paragraph{Experiments on NL2LF}
\label{sec:exp2}
Subsequently, we would like to focus on learning a model that given a NL labeling description, outputs a Python LF. To this end, we assemble a manually annotated dataset of pairs of NL descriptions to LFs, dubbed NL2LF in the following. To construct this dataset, we have scraped GitHub repositories containing LFs, that use the Snorkel library \citep{ratner2020snorkel}. Next, we manually annotated the code snippets, providing multiple descriptions for each LF, resulting in a dataset of 193 NL-LFs pairs.

In Table \ref{tbl:nl2lf} we report the results on experiments on the NL2LF dataset. For training, we use the same parameters as the ones for CoNaLa and we follow two different regimes: a) finetuning T5 models solely on the NL2LF training split, b) finetuning on the NL2LF training split, interleaved with batches from the gold CoNaLa dataset. A couple of interesting observations arise from this experiment: first, even if the NL2LF dataset is particularly small all T5 models manage to achieve a BLEU score of at least 0.30, a result illustrating the strength of the contextual representations learned by T5 during pre-training. These results are further improved when interleaving minibatches from NL2LF with the ones from the CoNaLa benchmark, achieving a BLEU of almost 0.39. Overall, these results demonstrate a promising direction towards using T5 to code NL descriptions to LFs, especially if we consider that the T5 models have not been pre-trained on specific text-code data and the fact that the NL2LF dataset is of particularly small.
\begin{table}
\caption{\label{tbl:nl2lf} Results on the NL2LF dataset with 5-fold cross validation 50-50 train-test splits.}
\centering
\begin{tabular}{ccc}
\hline 
Model&BLEU&Accuracy \\ 
\hline
\hline
T5-small on NL2LF&29.79&2.10\\
T5-base on NL2LF&30.00&3.09\\
T5-large on NL2LF&33.65&4.12\\
T5-small on CoNaLa+NL2LF&34.68&3.09\\
T5-base on CoNaLa+NL2LF&36.61&3.92\\
T5-large on CoNaLa+NL2LF&\textbf{38.97}&\textbf{5.15}\\
\hline 
\hline 
\end{tabular}
\end{table}

\section{Conclusions}
\label{sec:conclusions}
In this work we have focused on the task of converting natural language explanations to labeling functions through semantic parsing. We have presented a novel approach to dealing with the task of semantic parsing, employing deep pre-trained text-to-text Transformers (T5). Our approach achieves new state of the art on the CoNaLa benchmark. Additionally, we constructed manually a dataset with natural language explanation--labeling function pairs. Our best T5 model achieves a BLEU of 0.39 on our dataset, paving the way for practical applications of our approach in maximizing data labeling efficiency.
\section*{Broader Impact}

We believe that this work can have a broader impact on how humans annotate data for machine learning algorithms. Sourcing weak labeled data can become easier and without requiring programming knowledge or understanding on how machine learning models work. Broadly, we see this work as one more step towards teaching models \emph{how} to learn rather than providing them with direct annotations, leveraging both the domain expertise from annotators, rendering their job less repetitive and more efficient and eventually allowing an interface between humans and models in transferring labeling knowledge.

\bibliographystyle{acl_natbib}
\bibliography{main}

\end{document}